\documentclass[10pt,twocolumn,letterpaper]{article}

\usepackage{wacv}
\usepackage{times}
\usepackage{epsfig}
\usepackage{graphicx}
\usepackage{amsmath}
\usepackage{amssymb}

\usepackage{xr}
\graphicspath{{\string~/Box/L1_stabilization_paper_internal/}}

\def \R {\mathbb{R}}

\def\<{\langle}
\def\>{\rangle}

\usepackage{subfig}
\usepackage{cite}
\def\ie{\emph{i.e. }}
\def\etal{\emph{et al. }}
\usepackage[export]{adjustbox}

\def\wacvPaperID{515} 

\wacvfinalcopy 

\ifwacvfinal
\usepackage[breaklinks=true,bookmarks=false]{hyperref}
\else
\usepackage[pagebackref=true,breaklinks=true,colorlinks,bookmarks=false]{hyperref}
\fi

\begin{document}

\title{Cinematic-L1 Video Stabilization with a Log-Homography Model}

\author{Arwen Bradley, Jason Klivington, Joseph Triscari, Rudolph van der Merwe \\
Apple, Inc. \\
{\tt\small \{arwen\_bradley, klivington, jtriscari, vandermerwe\}@apple.com} \\
}

\maketitle
\thispagestyle{empty}

\begin{abstract}
We present a method for stabilizing handheld video that simulates the camera motions cinematographers achieve with equipment like tripods, dollies, and Steadicams. We formulate a constrained convex optimization problem minimizing the $\ell_1$-norm of the first three derivatives of the stabilized motion. 
Our approach extends the work of Grundmann \etal \cite{grundmann2011auto} by solving with full homographies (rather than affinities) in order to correct perspective, preserving linearity by working in log-homography space.
We also construct crop constraints that preserve field-of-view;
model the problem as a quadratic (rather than linear) program to allow for an $\ell_2$ term encouraging fidelity to the original trajectory; 
and add constraints and objectives to reduce distortion.
Furthermore, we propose new methods for handling salient objects via both inclusion constraints and centering objectives. 
Finally, we describe a windowing strategy to approximate the solution in linear time and bounded memory.
Our method is computationally efficient, running at 300fps on an iPhone XS, and yields high-quality results, as we demonstrate with a collection of stabilized videos, quantitative and qualitative comparisons to \cite{grundmann2011auto} and other methods, and an ablation study.
\end{abstract}

\vspace{-0.5cm}
\section{Introduction}

In recent years, individuals have increasingly been shooting video with smartphones and other handheld devices. Unlike professional cinematographers, however, they usually do not carry stabilization equipment such as tripods, dollies, or Steadicams, and they may not have the time or training to shoot with cinematic intent. We want to stabilize these casually-shot, handheld videos to clean up the shaky camera motion into the smooth motions characteristic of professional cinematography to create a more aesthetically pleasing video that allows the viewer to better focus on the content. Further, we may wish to guide the stabilization based on salient objects, for example to try to keep them centered or at least within the stabilized frame.
\begin{figure}
\begin{center}
\includegraphics[width=1.0\linewidth]{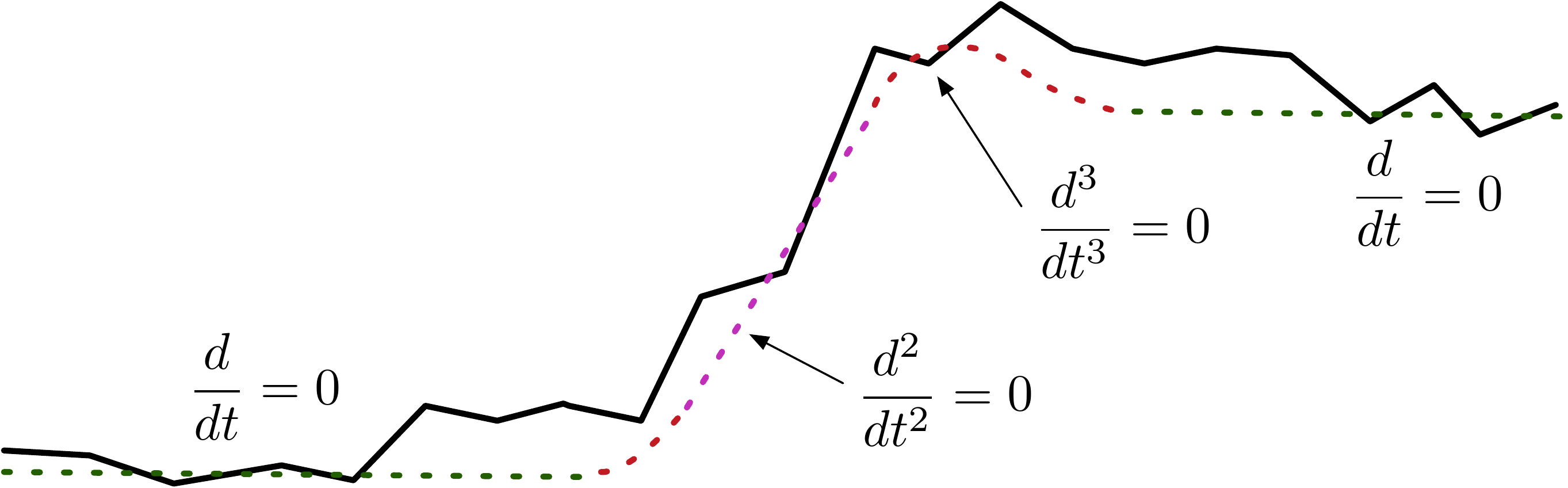}
\caption{We seek a stabilized trajectory that mimics the camera motions of professionally-shot video, namely stationary, panning, and constant acceleration motions that might be shot with a tripod, dolly, or Steadicam. These motions correspond to zero first, second, and third derivatives.}
\label{fig:sparse_deriv}
\end{center}
\end{figure}

\begin{figure}
\begin{center}
\includegraphics[width=1.0\linewidth]{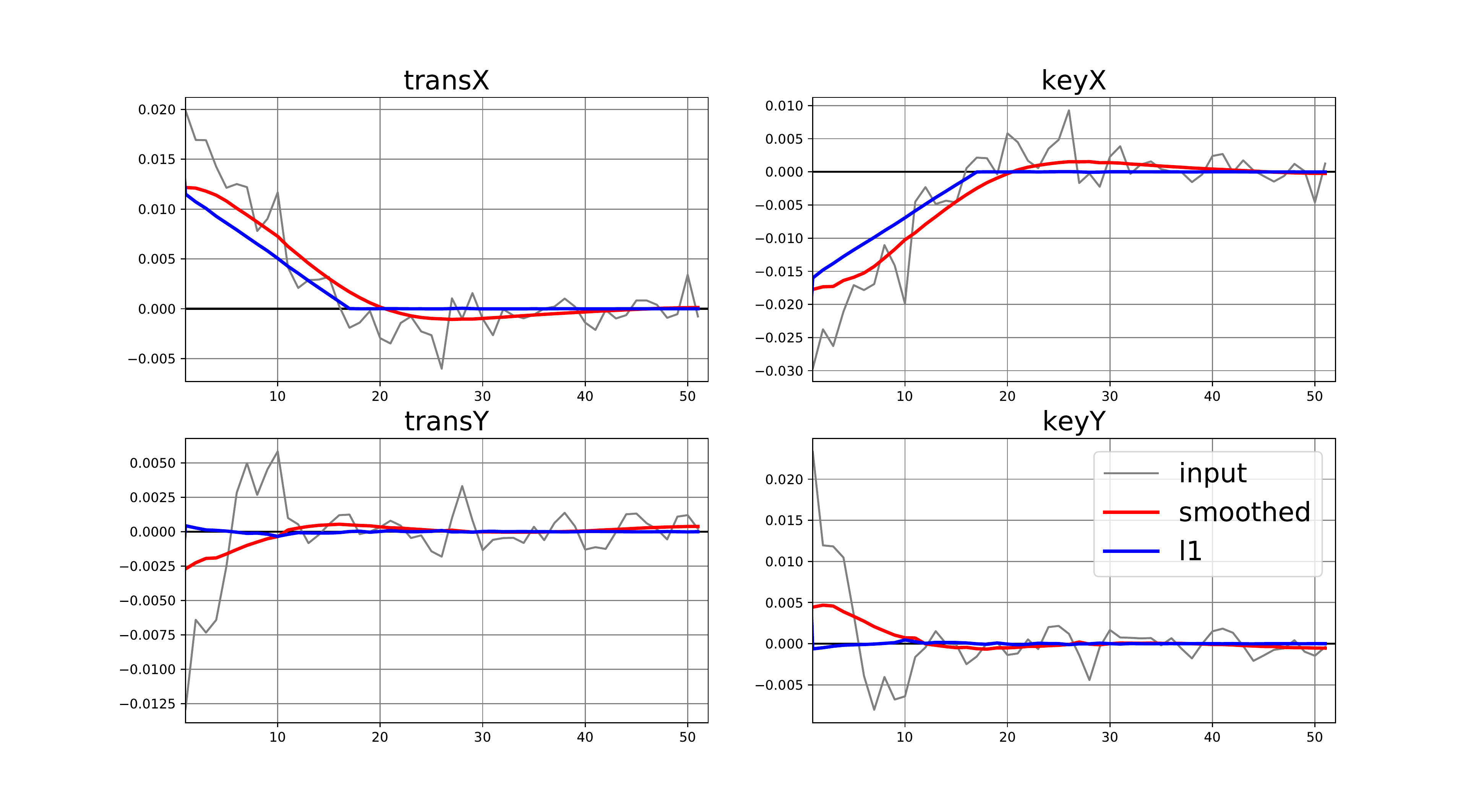} 
\caption{Comparison of Cinematic-L1 and Gaussian smoothing (corresponding to video \emph{fountain-tripod-smooth-l1.mp4} in supplementary material). Each subplot shows the value over time of a single translation or keystone homography element (where homographies map between consecutive frames). Gaussian smoothing just removes high-frequency motion, while L1 stabilization yields a constant-velocity pan followed by a `tripod' segment. (For corresponding derivative plots, see supplement.)}
\label{fig:fountain-l1}
\end{center}
\end{figure}

Video stabilization methods can generally be categorized as 2D, 2.5D, or 3D. 3D methods model the full 3D scene structure, often via Structure from Motion (SFM)~\cite{kopf2014first, ullman1979interpretation}, and stabilize using full 3D reprojections or intelligent approximations like Content Preserving Warps \cite{liu2009content}. 3D methods are the most powerful, but tend to be slower (SFM performance is improving but still costly) and less robust than simpler methods. 2.5D methods seek to handle depth variation in the scene without full 3D scene reconstruction and tend to be faster than 3D: \cite{liu2011subspace} exploit the fact that motion trajectories of rigid scenes tend to lie in low-dimension subspaces; \cite{liu2013bundled} uses multiple bundled homographies to handle scene depth variations; \cite{liu2014steadyflow, liu2016meshflow} perform spatial smoothing of an optical flow field and nonparametric mesh, respectively; and \cite{goldstein2012video} use an epipolar geometry model. 

Our approach belongs to the 2D class, which do not handle scene depth variation and can suffer from parallax, but have advantages of robustness and speed. In performance-critical contexts like capture-time video stabilization where inertial sensor (gyro/accel) data is available but pixel-based analysis is too costly, the scene structure cannot be determined and 2D models are the only choice. An important class of 2D methods focus on rolling-shutter correction~\cite{hanning2011stabilizing, ringaby2012efficient, grundmann2012calibration}; such methods are complementary to ours, and can be run as a preprocessing step (see supplement).

The basic stages of video stabilization are `analysis': determining frame-to-frame transforms that describe the original camera trajectory, and `correction': computing transforms to apply to each frame plus a final crop rectangle to generate a stabilized video. 2D analysis generally relies on either pixel-based image registration techniques~\cite{BattiatoEtAl2007}, or leveraging hardware motion sensor data~\cite{KarpenkoEtAl-CSTR-2011-03}. There are many possible strategies for the correction phase. One is `virtual tripod': mapping each frame back to a reference by chaining and inverting the analysis transforms; but this will fail for videos with too much motion (crop will be empty). Another is `smoothing' by applying a fixed or variable kernel (Gaussian, for example) to the input trajectory to remove the high-frequency motion \cite{hanning2011stabilizing}; this is fairly robust, though it never achieves perfect tripod-stabilization or constant-velocity pan. Neither tripod nor fixed-kernel smoothing offer control over the amount of crop, or the possibility of guidance based on saliency.

Inspired by the proposal of Grundmann \etal in~\cite{grundmann2011auto}, we focus on a powerful, flexible `L1' correction strategy. The goal is to create a stabilized trajectory consisting of stationary, constant-velocity pan, and constant-acceleration segments, so the video looks as though it had been shot with professional camera equipment (tripod, dolly, Steadicam, respectively). (This type of cinematography-inspired stabilization criteria was first proposed by~\cite{gleicher2008re}.)~\cite{grundmann2011auto} formulate the video stabilization problem as a linear program, use the $\ell_1$-norm heuristic to encourage sparsity~\cite{tibshirani94regressionshrinkage} in the derivatives of the smoothed trajectory (see figure \ref{fig:sparse_deriv}), and add constraints for crop, distortion, and possibly salient points. The convex formulation is key to good performance, since it allows for the use of interior-point methods, which efficiently solve large, sparse, convex problems~\cite{boyd2004convex}. Subsequently, convex `L1' type approaches have been explored by others:~\cite{gandhi2014multi} create multiple clips focusing on different actors with a convex formulation involving saliency constraints and an L1 penalty,~\cite{achary2019cinefilter} propose a windowing approach that allows for online processing, and~\cite{qu2013l1l2} add an $\ell_2$ term encouraging closeness to the input.

~\cite{grundmann2011auto}, as well as~\cite{gandhi2014multi, achary2019cinefilter, qu2013l1l2}, formulate the optimization problem using an affine model for the analysis and correction transforms. Affinities can capture translation, rotation, scale, and shear, but cannot model perspective (keystone), which limits the quality of stabilization. The more general class of homographies accounts for perspective, and fully captures the projective relationship between two images of a plane~\cite{hartley2003multiple}. Homographies are represented as 3x3 nonsingular \emph{homogeneous} matrices (that is, only the ratio of the elements is significant, or put another way, homographies are equivalent up to normalization). Homographies are a powerful tool for stabilization, but the nonlinearity poses many technical difficulties in the convex problem formulation. Although homographies themselves are convex~\cite{boyd2004convex}, compositions or other complex expressions involving homographies quickly become intractable.~\cite{grundmann2011auto} refine the affine solution with homographies in a post-process `residual motion suppression' step, which helps correct the residual keystone, but this is less principled than solving directly with homographies and can have undesirable side-effects like consuming significantly more crop.

\subsection{Contributions}

\begin{figure}
\begin{center}
\includegraphics[width=1.0\linewidth]{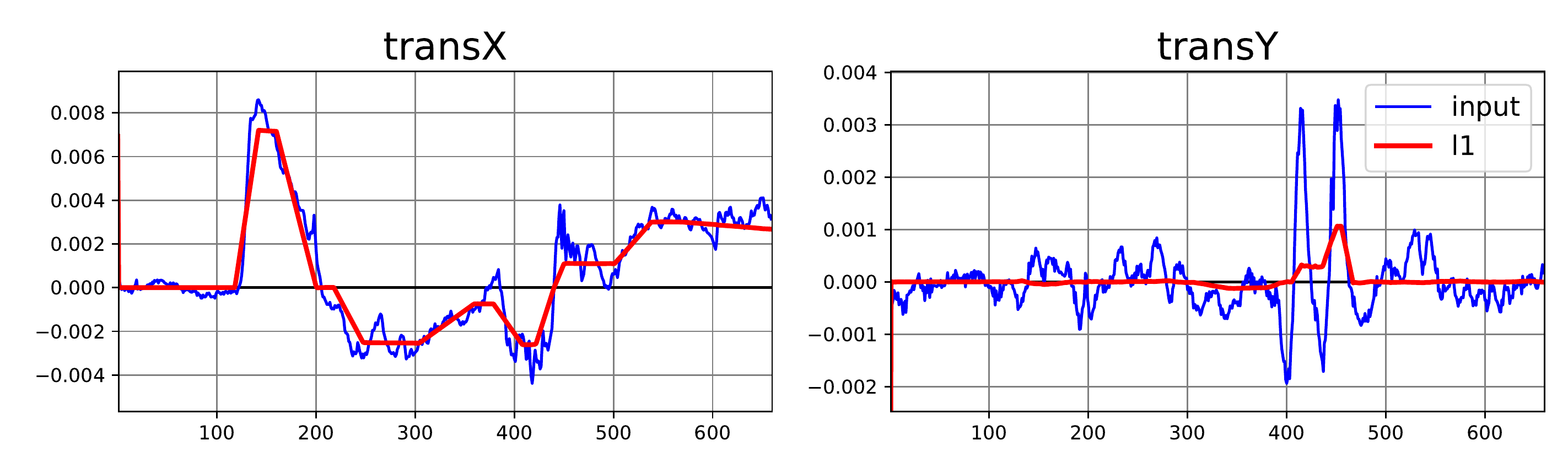}
\includegraphics[width=1.0\linewidth]{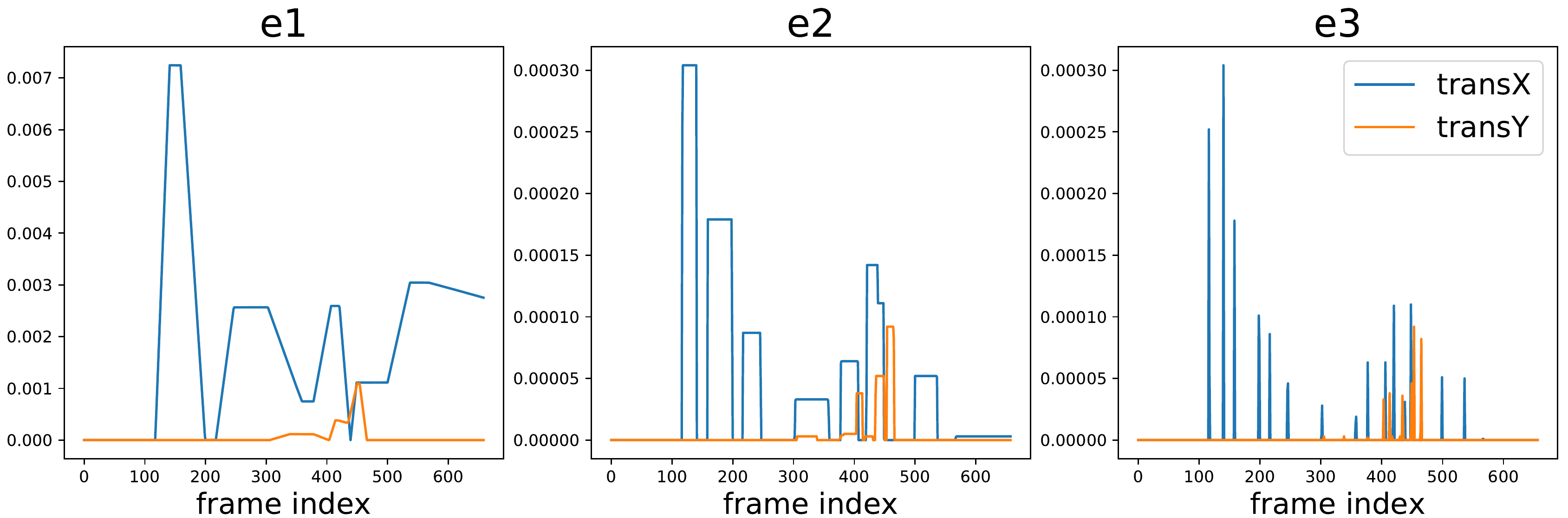}
\caption{L1 stabilization results for supplementary video \emph{river-l1.mp4}. \textbf{Top}: Translation elements of homographies over time (where homographies map between consecutive frames). \textbf{Bottom}: Sparse absolute values of first ($e^1$), second ($e^2$), and third ($e^3$) derivatives of L1 stabilized path (per equation \ref{eq:l1_problem}) (only translation elements are shown; see supplement for plots of all elements.)}
\label{fig:river-l1}
\end{center}
\end{figure}

Our strategy is to follow~\cite{grundmann2011auto} in formulating the video stabilization problem as a convex optimization problem, wherein we encourage sparsity in the first three derivatives of the stabilized motion by minimizing the $\ell_1$-norm, subject to crop and other constraints. We build upon their approach by solving with homographies rather than affinities (by introducing a log-homography transform), modeling the problem as a quadratic rather than linear program so we can add $\ell_2$ terms to discourage large excursions from the input trajectory, improving upon the crop constraints, addressing various distortion concerns, in particular some that arise from homographies as opposed to affinities, and proposing new ways to deal with salient objects with both constraints and centering objectives. Finally, we describe a windowing strategy to approximate the global L1 solution in linear time and bounded memory. Figure \ref{fig:fountain-l1} compares L1 stabilization to Gaussian smoothing. Figure \ref{fig:river-l1} shows L1 stabilization results for a longer, more complex, video.

\begin{itemize}
\itemsep-0.5em

\item \emph{Log-homography transform.}
Using homographies rather than affinities allows for much more precise, high-quality stabilization, but the nonlinearity introduced by the perspective transform creates serious technical difficulties in the convex formulation. In order to handle homographies while preserving convexity, we introduce a log-homography transform based on Lie theory that allows us to express the problem in terms of linear and quadratic operations.

\item \emph{Crop constraints preserving field-of-view.}
We develop crop constraints that not only ensure that the crop window samples valid pixels, but also that the transforms preserve field-of-view, via approximations in terms of area and sidelength. We show how to linearize these constraints using the Jacobian.
	
\item \emph{Fidelity objective term and variable stabilization strength.} We add an $\ell_2$ objective term to encourage fidelity to the input path in order to respect intent, reduce distortion, and optionally provide granular control over the stabilization strength. (This is similar to the $\ell_2$ term used in~\cite{qu2013l1l2}.)

\item \emph{Distortion mitigation.} Beyond the fidelity term, we also address distortion issues via constraints and objectives applied to the affine diagonal elements, off-diagonal elements, and homography keystone/translation ratios.

\item \emph{Saliency.} We propose two strategies for handling salient objects: constraints to keep salient objects from being cropped out of the stabilized frame, and centering objective terms to attempt to keep salient objects close to the center of the stabilized frame.

\item \emph{Windowed L1.} In order to handle arbitrary-length videos, we describe an overlapping-window strategy that runs in linear time and bounded memory and approximates the global solution given sufficient lookahead. A Markov property allows us to capture the complete history via a three-frame window initialization. (Both~\cite{achary2019cinefilter} and~\cite{qu2013l1l2} describe windowing approaches for affine models, but neither exploits the Markov property:~\cite{achary2019cinefilter}'s CineConvex formulation adds constraints on overlapping segments rather than just 3 frames, and~\cite{qu2013l1l2} compute independent solutions on overlapping windows and use a weighted average.)
\end{itemize}

\vspace{-0.1cm}
\section{L1 stabilization with log-homographies}
As discussed in the introduction, video stabilization consists of two phases, `analysis' and `correction', where the analysis phase seeks to estimate homographies mapping consecutive frames of the input, and the correction phase determines correction transforms to be applied to each frame, along with an overall crop. The focus of this work is on the second phase, so we assume from now on that the analysis homographies are known constants, denoted by $\{F_t\}$.
Let $\{P_t\}$ denote the homography corrections to be applied to a fixed centered \emph{crop window} to stabilize the video. (Note that we would apply $\{P_t^{-1}\}$ to the original video frames, then crop, to render the stabilized video.)
\begin{align}
F_t &\quad \text{maps input $t \to t-1$ } \notag \\
\tilde{F}_{t} &= F_0 \ldots F_{t} \quad \text{maps input $t \to 0$} \notag \\
S_t &= \tilde{F}_{t} P_t \quad \text{maps stabilized $t \to 0$}
\label{eq:F_t_P_t_S_t}
\end{align}
We formulate the stabilization problem as a convex optimization problem in the variables $P_t$, with constraints and objectives that achieve the following goals:
\begin{itemize}
\itemsep-0.5em
\item encourage sparse first, second, and third derivatives of the stabilized motion $S_t$ by minimizing the $\ell_1$-norm (as proposed by~\cite{grundmann2011auto}) -- these zero derivatives correspond to tripod, panning, and smoothly accelerating motions.
\item encourage fidelity of the stabilized to the original camera path, by keeping corrections small in an $\ell_2$ sense.
\item respect crop constraints by ensuring both valid pixels in the crop window and preservation of field-of-view.
\item minimize distortion via objective terms and constraints that control the relationships between elements of the correction homographies.
\item optionally, keep `salient' objects either approximately centered or in frame.
\end{itemize}

In order to support homographies (as opposed to affinities) while maintaining convexity, we linearize the problem by transforming to the vector space of log-homographies. In this section, we describe the log-homography transformation (with a brief introduction to the Lie theory that supports it) and show how to express or approximate each of the items above in the resulting framework.

\subsection{Log homographies}

\begin{figure}
\begin{center}
\includegraphics[width=0.7\linewidth]{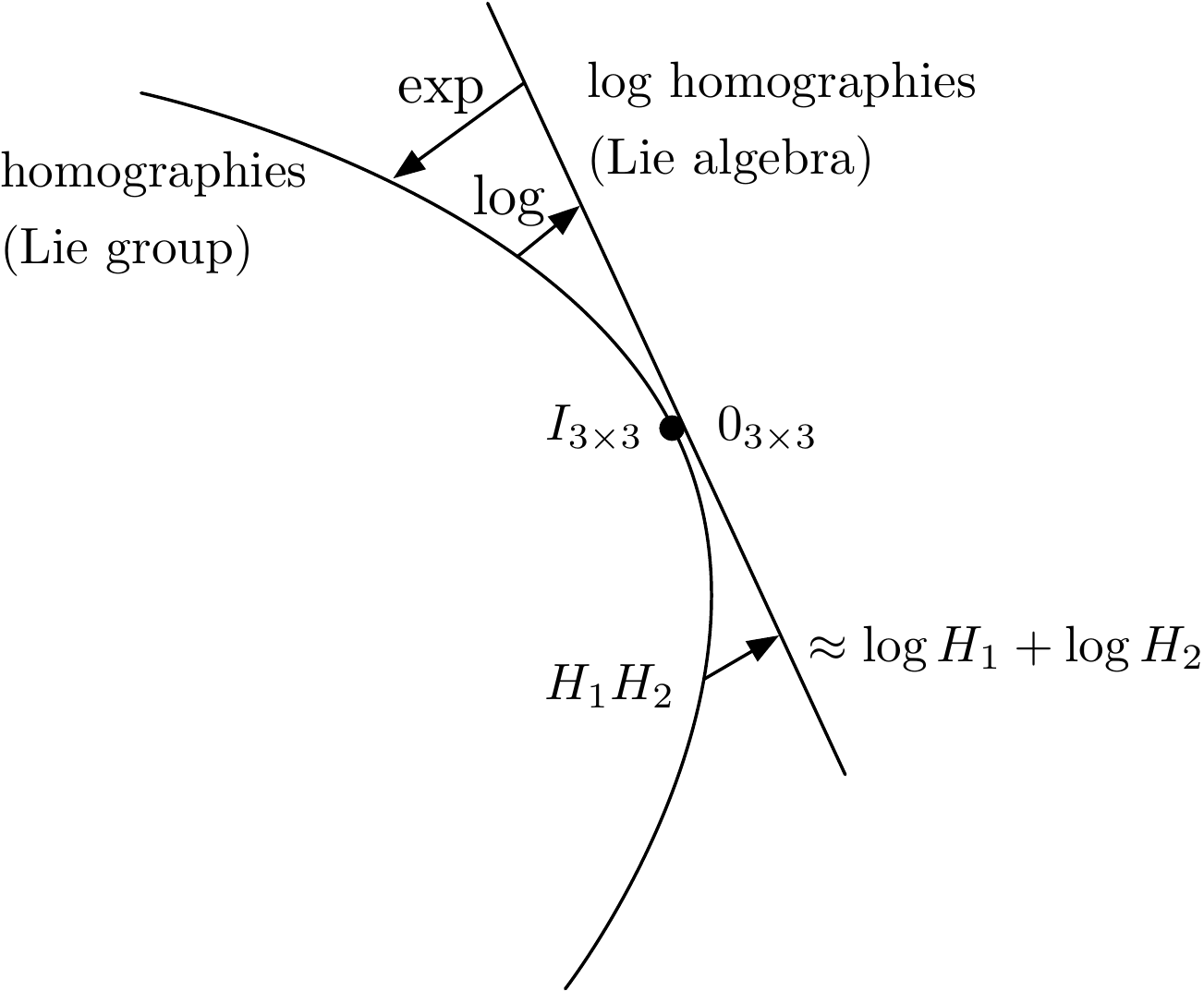}
\caption{The space of homographies is a Lie group, and its tangent space at the identity forms a Lie algebra (a vector space). The matrix exponential maps from the Lie algebra to the Lie group; its inverse is the matrix logarithm. We will formulate the L1 stabilization problem in terms of log-homographies (elements of the Lie algebra).}
\label{fig:homogmanifold}
\end{center}
\end{figure}

The basic motivation for the log-homography transform is to turn complex nonlinear operations (composition and inverses of homographies) into simple linear operations (addition and negation of log-homographies, respectively). These properties are analogous to those of the standard logarithm on real numbers:
$ \log(xy) = \log(x) + \log(y)$, $\log(x^{-1}) = -\log(x)$.
Lie theory allows us to formalize this notion. We will briefly outline the key points relevant to our application here; for more detail see for example~\cite{lee2013smooth}. 
The group of homographies is isomorphic to  $SL(3, \R)$, the group of real invertible square matrices with determinant 1, since homography matrices differing only by a scale factor are equivalent and we can normalize to determinant 1. $SL(n, \R)$ is a Lie group, which is both a group and a manifold. As a manifold, $SL(n, \R)$ is a complex surface where addition is undefined (think of the surface of a sphere), but by working in the tangent space we can recover the notion of addition. In general, the tangent space at the identity of a Lie group forms a Lie algebra (see figure \ref{fig:homogmanifold}), which unlike the group itself is a vector space: in particular, we have addition and scalar multiplication. The Lie algebra of $SL(n, \R)$, denoted  $\mathfrak{sl} (3, \Re)$, consists of $3 \times 3$ matrices with zero trace; the identity of $SL(n, \R)$ is the usual identity matrix $I$ and the identity of $\mathfrak{sl} (3, \Re)$ is the zero matrix. The exponential map from the Lie algebra to the manifold is given by the matrix exponential $\exp(X) = \sum_{k=0}^\infty \frac{X^k}{k!}$. (This exponential is related to but different from the Riemannian exponential used for example in \cite{zhang2017geodesic} for geodesic interpolation between key frames.) In a neighborhood of the identity, the exponential map has an inverse: the matrix logarithm. That is, for a homography $H$:
\begin{align}
H = \exp (h) \in SL(3, \R), \quad
h = \log(H) \in \mathfrak{sl} (3, \Re). 
\end{align}
There are a variety of methods for computing exp and log available in many scientific computing libraries. Our implementation perturbs the matrix until it is diagonalizable and exploits the fact that if $\text{exp}(A) = Ue^DU^{-1}$ with $D = \text{diag}(d)$ then $e^D = \text{diag}(e^d)$. Other implementations use the Pade approximation to the infinite series. Two important properties for our application are that, to first-order in a neighborhood of the identity, we have:
\begin{align}
\notag \log (H^{-1}) &\approx -h; \\
\log (H ) &\approx \log(H_1) + \log(H_2).
\label{eq:log_sum_inv}
\end{align}
Both properties follow from the fact that group multiplication in a Lie group is reflected to first order in the vector space structure of its Lie algebra ~\cite{lee2013smooth}. Neither can hold exactly since homographies do not commute in general, so these are approximations, but nevertheless useful. The linearity of log-homographies enables us to formulate L1 stabilization as a convex problem, which would be difficult or impossible with homographies.

\subsection{L1 stabilization with log-homographies}


Now that we have the log-homography transformation tools, we show how to formulate the stabilization problem. With the notation of equation \ref{eq:F_t_P_t_S_t},  let $f_t = \log F_t$ denote the input log-homographies. The problem variables are the correction log-homographies $p_t = \log P_t \in \R^9$, which we stack into a column vector $p \in R^{9n}$ where $n$ is the number of frames. We also introduce auxiliary variables $e_1 \in \R^{9(n-1)}, e_2 \in \R^{9(n-2)}, e_3 \in \R^{9(n-3)}$ representing the first, second, and third derivatives of the smoothed path. The constants $w_0, w_1, w_2, w_3$ are fixed weights.

We define the optimization problem:
\begin{align}
\underset{p \in \R^{9n} }{\text{min}} \quad & \quad  \frac{1}{2}w_0 \| p \|_2^2 + w_1 \|e^1\|_1 + w_2 \|e^2\|_1 + w_3 \|e^3\|_1 \notag \\
\text{s.t.:}
& \quad \text{lb} \le p_t \le \text{ub}, \quad \forall t \notag \\
& \quad \text{trace}(p_t) = 0, \quad \forall t \notag \\
&\quad \text{linear crop constraints on $p_t$}, \quad \forall t \notag \\
\text{where  }\notag 
p &= (p_1, \ldots, p_n)^T, \quad p_t \in \R^9 \notag \\ 
e^1(t) &= p_{t+1} + f_{t+1} - p_t \notag \\
e^2(t) &= p_{t+2} + f_{t+2} - 2p_{t+1} - f_{t+1} + p_t \notag \\
e^3(t) &= p_{t+3} + f_{t+3} - 3p_{t+2} \notag \\
& \quad - 2f_{t+2} + 3p_{t+1} + f_{t+1} - p_t
\label{eq:l1_problem}
\end{align}

\begin{figure}
\begin{center}
\includegraphics[width=0.8\linewidth]{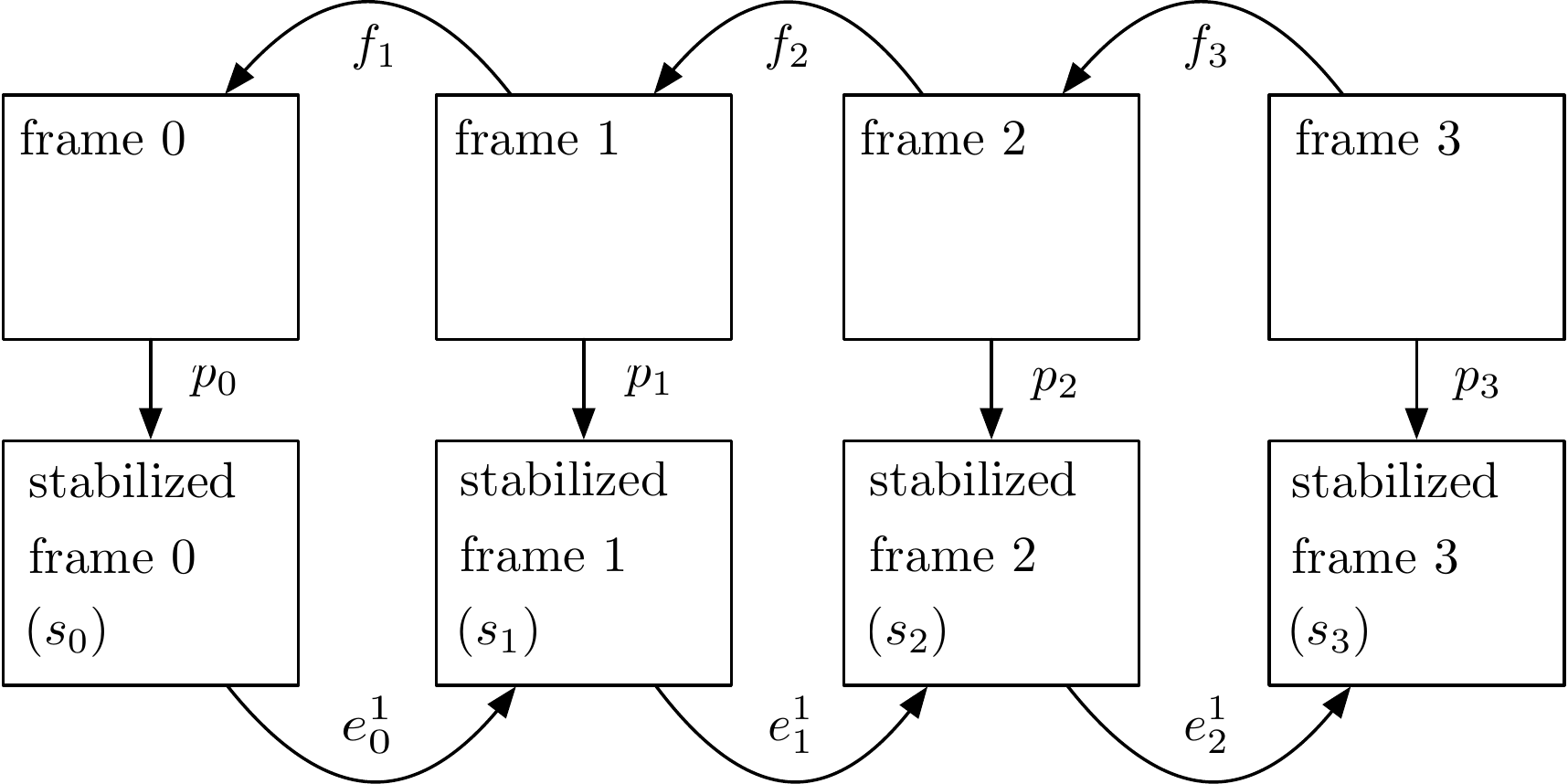}
\caption{The analysis log-homographies $f_t$ map consecutive input frames. The corrections $p_t$ are applied to a fixed crop window centered in each frame to stabilize the video. The derivatives $e^i$, $i=1,2,3$, refer to the stabilized path.}
\label{fig:corrections_derivs}
\end{center}
\end{figure}

The objective term $\frac{1}{2}w_0 \| p \|_2^2$ is intended to keep the correction log-homographies close to the identity, the goal being a stabilized path that is fairly faithful to the input path (see subsection \ref{section:fidelity}). The objective term $w_1 \|e^1\|_1$, along with the corresponding constraint defining $e^1$, is intended to promote sparsity of the first derivative of the smoothed path via the $\ell_1$-norm sparsity heuristic. $e^1$ is the first derivative of the stabilized log-path, $\tilde s_t = \tilde f_t + p_t$, so, using forward-differences:
\begin{align*}
e_1(t) &= \tilde s_{t+1} - \tilde s_t 
= \tilde f_{t+1} + p_{t+1} - \tilde f_t - p_t \\ 
&= p_{t+1} + f_{t+1} - p_t \quad \text{(since $\tilde f_{t+1} - \tilde f_t = f_{t+1}$)}.
\end{align*}
The $e^2$ and $e^3$ terms are similar, corresponding to second and third derivatives of the smoothed path (see supplement). The trace constraint is needed because log-homographies have zero trace, and the crop constraints are described next.

\subsection{Crop constraints}


\begin{figure}
\begin{center}
\includegraphics[width=1.0\linewidth]{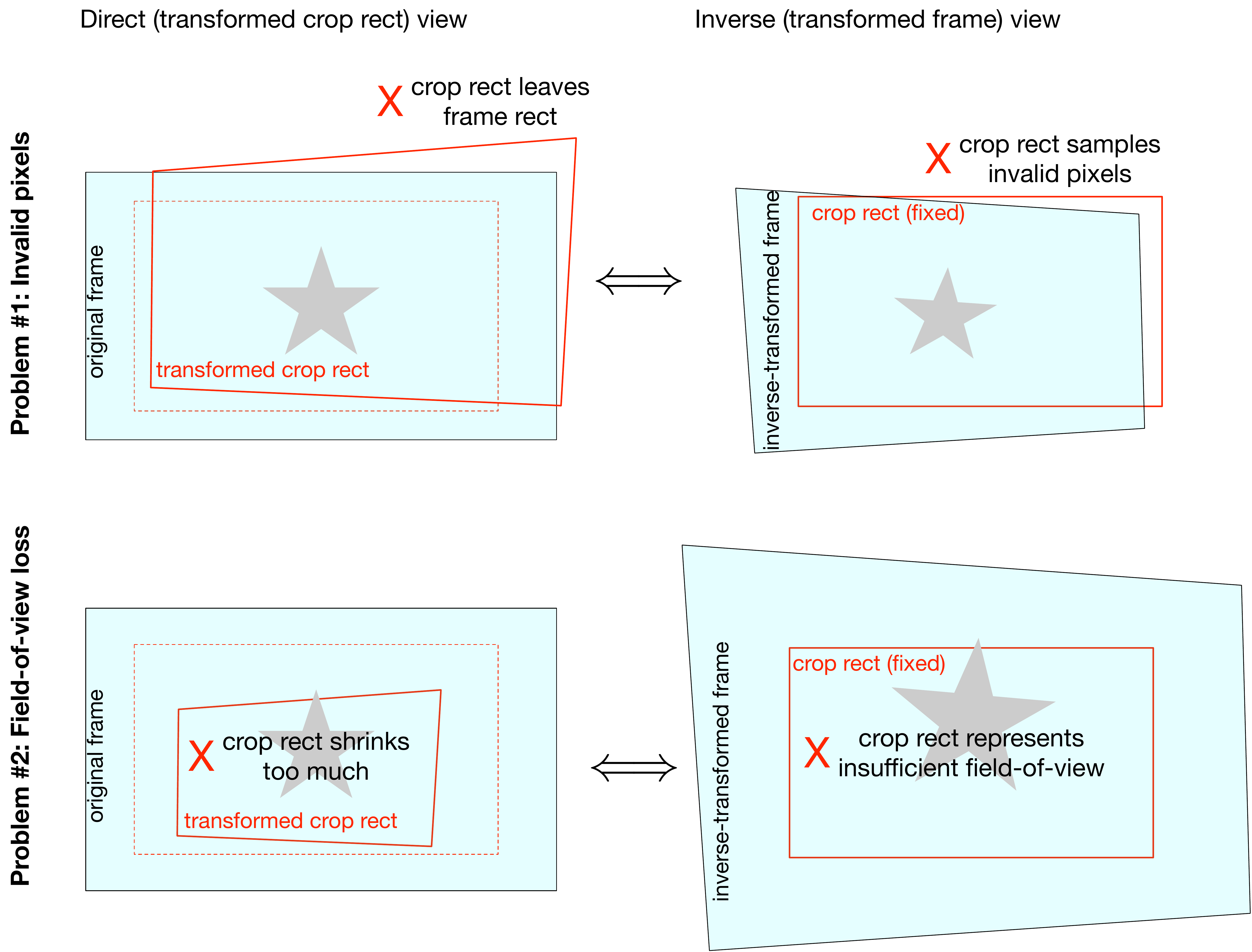}
\caption{There are two basic problems that our crop constraints must guard against: sampling invalid pixels and loss of field-of-view. Recall that there are two equivalent views of the problem: corrections $\{P_t\}$ applied to the crop window within a fixed frame, or inverse-transforms $\{P_t^{-1}\}$ applied to the original frame, viewed through a fixed crop window. Here we show the two views on the two problems. Problem 1: the transformed crop rect leaves the frame (equivalently, the fixed crop rect samples from outside the inverse-transformed frame), and hence contains invalid pixels. Problem 2: the transform shrinks the crop rect too much (equivalently, inverse transform enlarges the frame too much), so that the crop represents insufficient field-of-view (despite a fixed size in pixels).}
\label{fig:crop_problems}
\end{center}
\end{figure}

Stabilization necessarily results in crop (based on the intersection of valid pixels after all frames have been transformed by their respective corrections); in general, there is a tradeoff between stabilization strength and crop severity. We provide for adjustment of this tradeoff via a maximum-crop-allowance parameter, and build crop constraints to ensure that the stabilization does not exceed this allowance. A somewhat subtle point is that the crop fraction doesn't refer directly to the \emph{dimensions} of the stabilized and cropped video, but to the fraction of the \emph{field-of-view} represented therein. In other words, the maximum-crop parameter is a lower bound for the preserved field-of-view.

We model the problem in terms of the corrections to be applied to the crop window (which we generally make somewhat larger than the minimum allowed dimensions for the client-specified crop fraction, since the correction transformations may reduce its field-of-view). As shown in figure \ref{fig:crop_problems}, we need to protect against two basic problems: invalid pixels and loss of field-of-view. Hence we design constraints that ensure that (1) the transformed crop window stays within the original frame rectangle, ensuring all pixels are valid, and (2) sufficient field-of-view is preserved after transforming the crop window.
\\
\begin{bf}Valid-pixel constraints\end{bf}
We want to ensure that the transformed crop window lands fully inside the frame rectangle, so that it samples only valid pixels (to avoid problem 1 in figure \ref{fig:crop_problems}). That is, if $D(h; c^{0:4}) : \R^9 \to \R^8$ computes the change in the positions of four corners due to a log-homography $h$, we want
\vspace{-0.2cm}
\[ (-\frac{w}{2}, -\frac{h}{2}) \le D(p_t; c_\text{wind}^{0:4})  + c_\text{wind}^{0:4} \le (\frac{w}{2}, \frac{h}{2}), \]
for each correction log-homography $p_t$, where $c_\text{wind}^{0:4}$ are the 4 corners of the original crop rectangle and $w,h$ are the width and height of the original frame, with origin at the center of the frame. We linearize the constraints using the approximation $D(p_t; c_\text{wind}^{0:4}) \approx \nabla D |_{0, c_\text{wind}^{0:4}} p_t$, where $\nabla D \in \R^{8 \times 9}$ is the Jacobian with respect to $h$ centered at $0$, which linearly approximates how the corners move due to $h$.
\\
\begin{bf}Field-of-view constraints\end{bf}
The correction transformation must also preserve sufficient field-of-view (\ie avoid problem 2 in figure \ref{fig:crop_problems}). Since we can't express that requirement as a linear constraint, we instead approximate it by requiring that the area and sidelengths of the transformed crop window are at least as large as the area and sidelengths of a rectangle with dimensions $(1 - \text{crop fraction})$ times the frame dimensions. These constraints do not necessarily guarantee field-of-view preservation; under extreme distortion -- large shear, for example -- they can be satisfied but field-of-view can still be lost. However, our distortion constraints (section \ref{section:distortion}) and objective terms generally prevent such pathological cases from occurring. So, the approximate field-of-view constraints are:
\vspace{-0.1cm}
\begin{align*}
\text{\small{Area}}(c(p; c_\text{wind})) &\ge (1 - \text{\small{crop frac}})^2 \times \text{\small{Area}}(\text{\small{frame}}) \\
\text{\small{Sidelen}}(c(p; c_\text{wind})) &\ge (1 - \text{\small{crop frac}}) \times \text{\small{Sidelen}}(\text{\small{frame}}),
\end{align*}
for each correction $p$, where $c(p; c_\text{wind})$ denotes the corners of the crop window transformed by $p$. As with the valid-pixel constraints, we can linearize the area and sidelength constraints via the Jacobian (details in supplement).

\subsection{$\ell_2$ Fidelity objective} \label{section:fidelity}

The quadratic term $\frac{1}{2}w_0 \| p \|_2^2$ helps keep the correction log-homographies close to the identity. This encourages fidelity to the input camera trajectory in order to respect the original intent. (For example, if the input is a deliberate pan, we should not necessarily tripod-stabilize even if the constraints would allow.) The $\ell_2$-term also acts as regularization, making the problem strictly convex and the solution unique. (Without saliency constraints, \cite{grundmann2011auto} is not strictly convex.) 
The fidelity term also offers control over stabilization strength, since varying $w_0$ trades off fidelity versus minimizing derivatives. $w_0$ can even be adjusted per-frame; for example, we may want to decrease stabilization strength for frames with serious motion blur, since removing the camera motion associated with motion blur can look unnatural.

\subsection{Distortion mitigation} \label{section:distortion}

We attempt to minimize the perceived distortion that can result from stabilization corrections. For example, shearing and perspective transforms can create the appearance of unnatural, fluid motion of rigid objects, and planar distortions of non-planar content like faces can produce unpleasant results, particularly if the distortions vary rapidly over time.
\\
\begin{bf}Affine diagonal and off-diagonal. \end{bf}
Grundmann~\cite{grundmann2011auto} encourage rigidity in the correction transforms by constraining the diagonal and off-diagonal absolute difference and sum \ie $|a-d| < \epsilon_1$, $|b+c| < \epsilon_2$, where $a,b,c,d$ are the entries of the upper left $2\times2$ block of the corrections. We find it more helpful to add quadratic objective terms $||a-d||^2$, $||b+c||^2$ penalizing these quantities -- the quadratic choice is mild for small values but becomes gradually severe for larger ones, as opposed to a constraint which provides no discouragement until hitting a hard stop.
\\
\begin{bf}Keystone-translation ratio. \end{bf}
The supplement shows that keystone and translation have a constant linear relationship under certain common circumstances. When this relationship exists in the input we should preserve it in the stabilized trajectory to avoid distortion. Given a keystone-translation ratio $R$ in the input: \ie $k \approx R t$, where $k$ is keystone and $t$ is translation, we can add an objective term: $\| k_x - R t_x \|^2  + \| k_y - R t_y \|^2$ to encourage the desired relationship in the corrections. If the input and correction transforms preserve the ratio, the stabilized path will also.

\vspace{-0.1cm}
\subsection{Saliency constraints and centering objective}

Many videos contain one or more salient objects or regions (people's faces, pets, or players at sports events \cite{chen2017should}, for example). We may wish to direct the stabilization path to `focus' on, or at least avoid cropping out, these objects. In this section we present a few approaches to this problem.
\\
\begin{bf}Inclusion constraints. \end{bf}
\begin{figure}
\begin{center}
\includegraphics[width=1.0\linewidth]{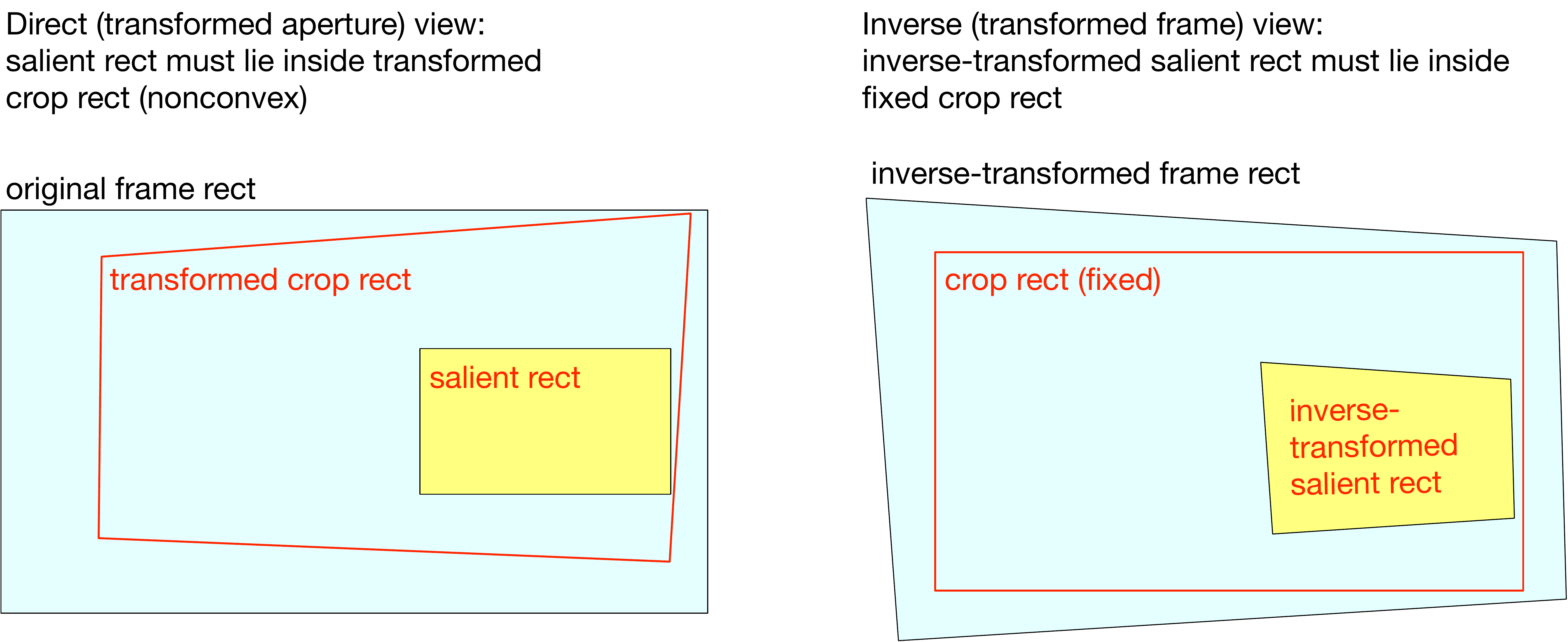}
\caption{To ensure that a salient rectangle remains visible in the stabilized video, we must keep it inside the transformed crop window, or equivalently, make sure that the inverse-transformed salient rect is contained in the fixed crop rect. We can approximate the latter viewpoint as a convex constraint in terms of log-homographies.}
\label{fig:saliency_constraints}
\end{center}
\end{figure}
\begin{figure}
\begin{center}
\includegraphics[width=1.0\linewidth]{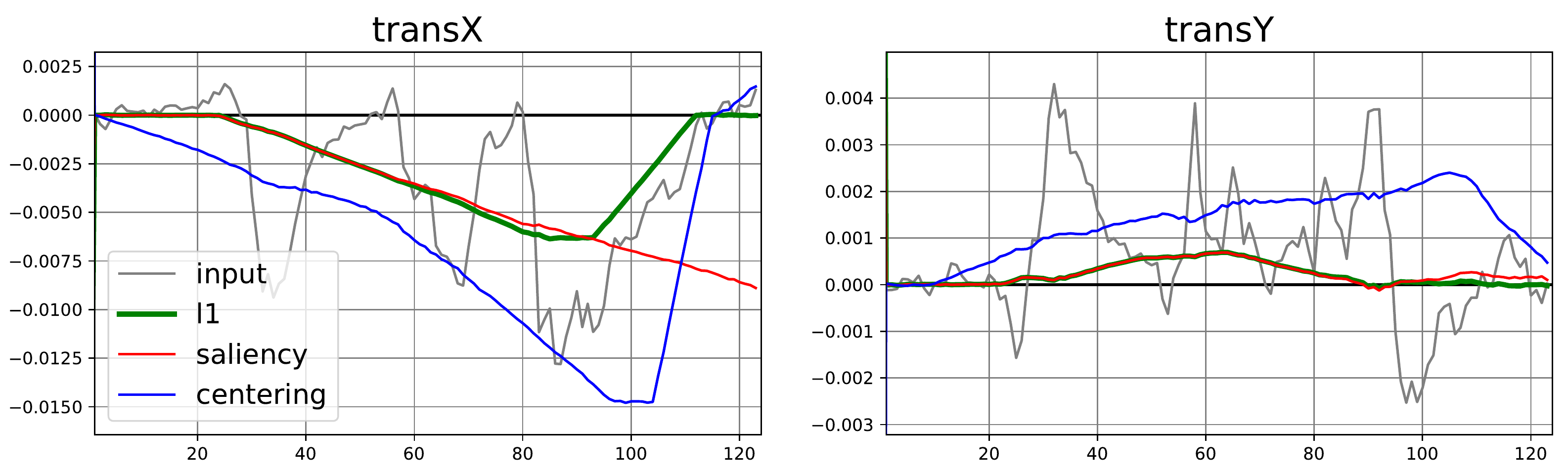}
\caption{Stabilization with saliency constraints or centering objective (please see supplementary video \emph{dog-l1-saliency-comparison.mp4}). Plots show value over time of translation homography elements (where homographies map between consecutive frames). Each plot compares default L1 (green), L1-with-saliency-constraint (red), and L1-with-centering-objective (blue).}
\label{fig:dog-saliency-homogs}
\end{center}
\end{figure}
One approach is to add hard or soft constraints requiring that salient points are included in the cropped, stabilized video. This is similar in spirit to the proposal of Grundmann, but our log-homography transform simplifies the formulation considerably. This approach can handle any collection of salient points, arbitrary salient regions (by computing the convex hull and placing a point at each vertex), and bounding boxes (by placing a point at each corner). Suppose we wish to ensure that a point $s_t \in \R^2$ remains visible in frame $t$ of the stabilized video. That is, $s_t$ should be inside in the crop window after it has been corrected by $P_t = \exp(p_t)$. We can't express this directly as a convex constraint, so we instead note that it is equivalent to $P_t^{-1} s_t \in \text{crop rect}$.
($P_t^{-1}$ is the correction that would be applied to the original frame to stabilize it.) Using our earlier notation and linearization strategy from the crop constraints, $d(h;x)$ represents the delta in point $x$ due to the corresponding homography $H = \exp(h)$, and so, using equation \ref{eq:log_sum_inv} ($\log H^{-1} = -h$), we have 
\vspace{-0.2cm}
\[ P_t^{-1} s_t = s_t + d(-p_t, s_t) \approx s_t - \nabla d|_{0, s_t} p_t ,\]
Hence we can approximate the constraint by the linearization:
$s_t - \nabla d|_{0, s_t} p_t \in \text{crop rect}$.
~\cite{grundmann2011auto} make a careful distinction between `feature path' and `camera path' formulations, (where corrections are applied to the video frame or crop window, respectively) -- the corrections in the two cases are inverses of each other. The former choice makes the saliency constraints easy to formulate but the crop constraints difficult, and they cannot express some constraints in terms of $P_t$ and others in terms of $P_t^{-1}$ since the inverse operation is non-convex. But log-homographies avoid that issue, since the inverse of $p_t$ is just $-p_t$.
\\
\begin{bf}Centering objective. \end{bf}
We can also add an objective term that attempts to center a salient object in the stabilized video, creating a stronger sense of `focus' than constraints alone. In general, we can guide the stabilized trajectory by adding an objective term to keep the corrections close to a sequence of target log-homographies: $\| p - p_\text{target} \|_2^2$ (we choose an $\ell_2$ penalty to strongly discourage large deviations while accepting small ones). In this case, we can set the target as the pure-translation corrections that would center the salient object in each frame. An advantage of objective terms versus constraints is that infeasibility is not an issue -- cases where the object cannot be centered are handled gracefully. Some disadvantages are that pulling toward targets far from the identity can introduce distortion, and such targets cannot be achieved without a large crop budget. For examples, please see figure \ref{fig:dog-saliency-homogs} and the referenced video.

\section{Windowed L1 stabilization} 

\begin{figure}
\begin{center}
\includegraphics[width=.7\linewidth]{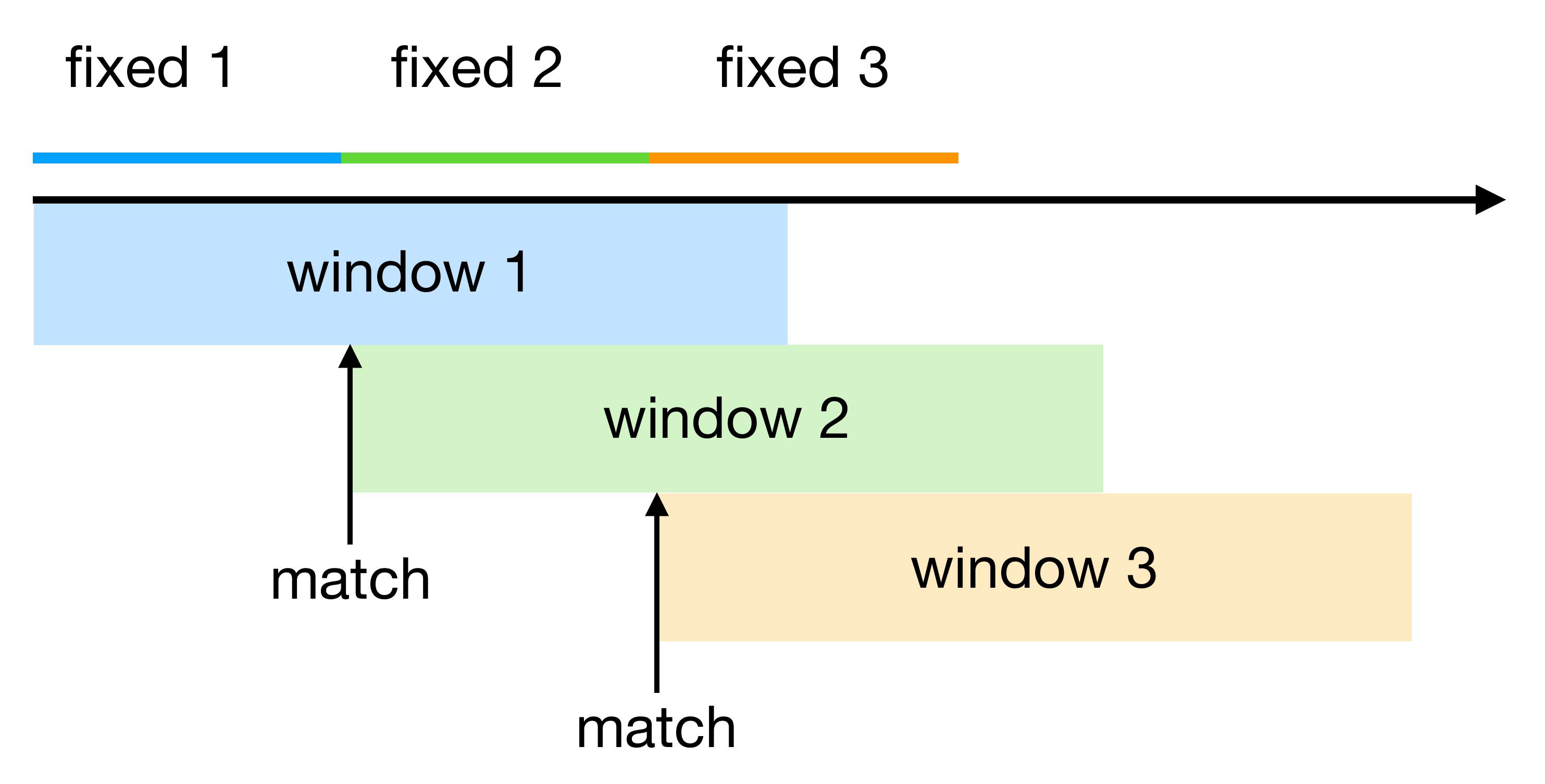}
\caption{Basic windowed L1 strategy: iteratively solve within overlapping windows subject to constraints requiring beginning of current window to match previous fixed solution.}
\label{fig:wind_l1}
\end{center}
\end{figure}

\begin{figure}
\begin{center}
\includegraphics[width=1.0\linewidth]{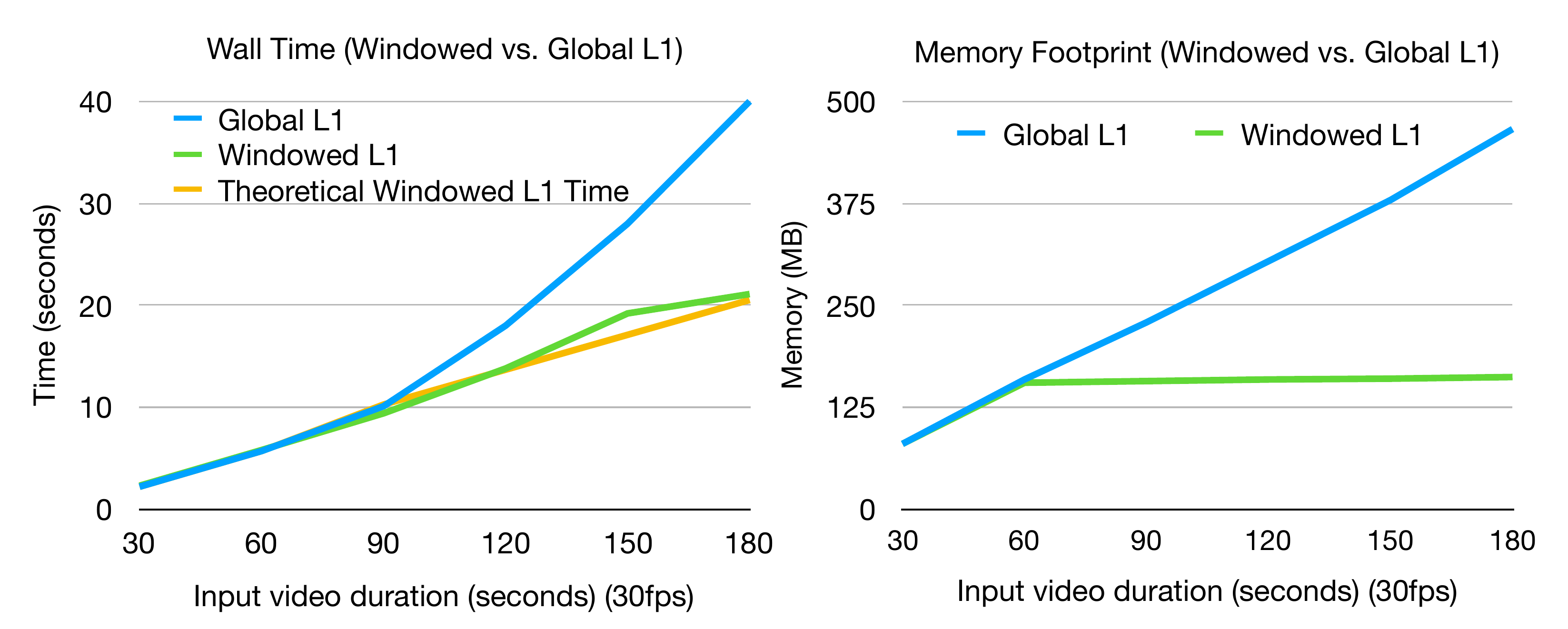}
\caption{Time and memory of global versus windowed L1 stabilization measured on an iPhone XS, using window length $l_w = 1800$ (= 60s@30fps) and stride $l_s = 1500$ (= 50s@30fps). Global L1 time grows superlinearly and memory grows linearly in the number of input frames, while windowed L1 time grows linearly and memory is bounded. The measured windowed L1 time agrees well with the theoretical time, which is $n \times \frac{\text{L1 time for $l_w$ frames}}{l_s}$.}
\label{fig:wind_l1_time_mem}
\end{center}
\end{figure}

To handle arbitrary-length videos, we describe a windowing strategy to approximate the solution in linear time and bounded memory. 
L1 stabilization models the problem as a sparse quadratic program that scales linearly with the frame count, so the memory footprint grows unboundedly as the frame count increases. Also, the processing time grows superlinearly in the number of frames (the majority of L1 time is spent in an interior-point solver bottlenecked by an LDL factorization; dense LDL is in $O(n^3)$, while sparse LDL depends on the sparsity pattern: for this problem, it is roughly quadratic). Therefore, in order to solve in linear time with bounded memory, we propose a windowed strategy. Consider a window length $l_w$ and stride length $l_s < l_w$ (in frames). As shown in figure \ref{fig:wind_l1}, the strategy is:
\begin{enumerate}
\itemsep0em
\item Set window start to zero: $s_w = 0$.
\item Solve global problem in window $[s_w, s_w + l_w)$.
\item \label{step:window_fix} Fix the solution for the first $l_s$ frames of the window.
\item Move window forward by $l_s-3$ frames: $s_w \mathrel{{+}{=}} l_s-3$.
\item \label{step:window_match} Solve global problem in window $[s_w, s_w + l_w )$ subject to initialization constraint forcing first 3 frames of current window to match previous fixed solution: $p_t = p^\text{fixed}_t$ for $s_w \le t < s_w+3$.
\item Repeat (\ref{step:window_fix})-(\ref{step:window_match}) until the current window reaches end of the video, truncating the final window as necessary.
\end{enumerate}
Step (\ref{step:window_match}) is the crux: with the solution fixed up to frame $s_w+3$, we actually wish to find the optimal solution in the entire range $[0, s_w + l_w )$ that agrees with the fixed solution in the range $[0, s_w+3)$. A third-order Markov property lets us achieve this by initializing only the first 3 frames of the current window. The intuition is that the past and future are linked only by derivatives up to third order, so the effect of all previous frames is only propagated through derivatives that depend on the previous 3 frames. Details are in supplement. With the windowed strategy, the memory footprint is bounded and equal to the memory required for a single window; the wall time for an input with $n > l_w$ frames is roughly $n \times \frac{\text{L1 time for $l_w$ frames}}{l_s}$. See figure \ref{fig:wind_l1_time_mem}.

\section{Results}

\begin{figure}[t]
\begin{center}
\includegraphics[width=1.0\linewidth]{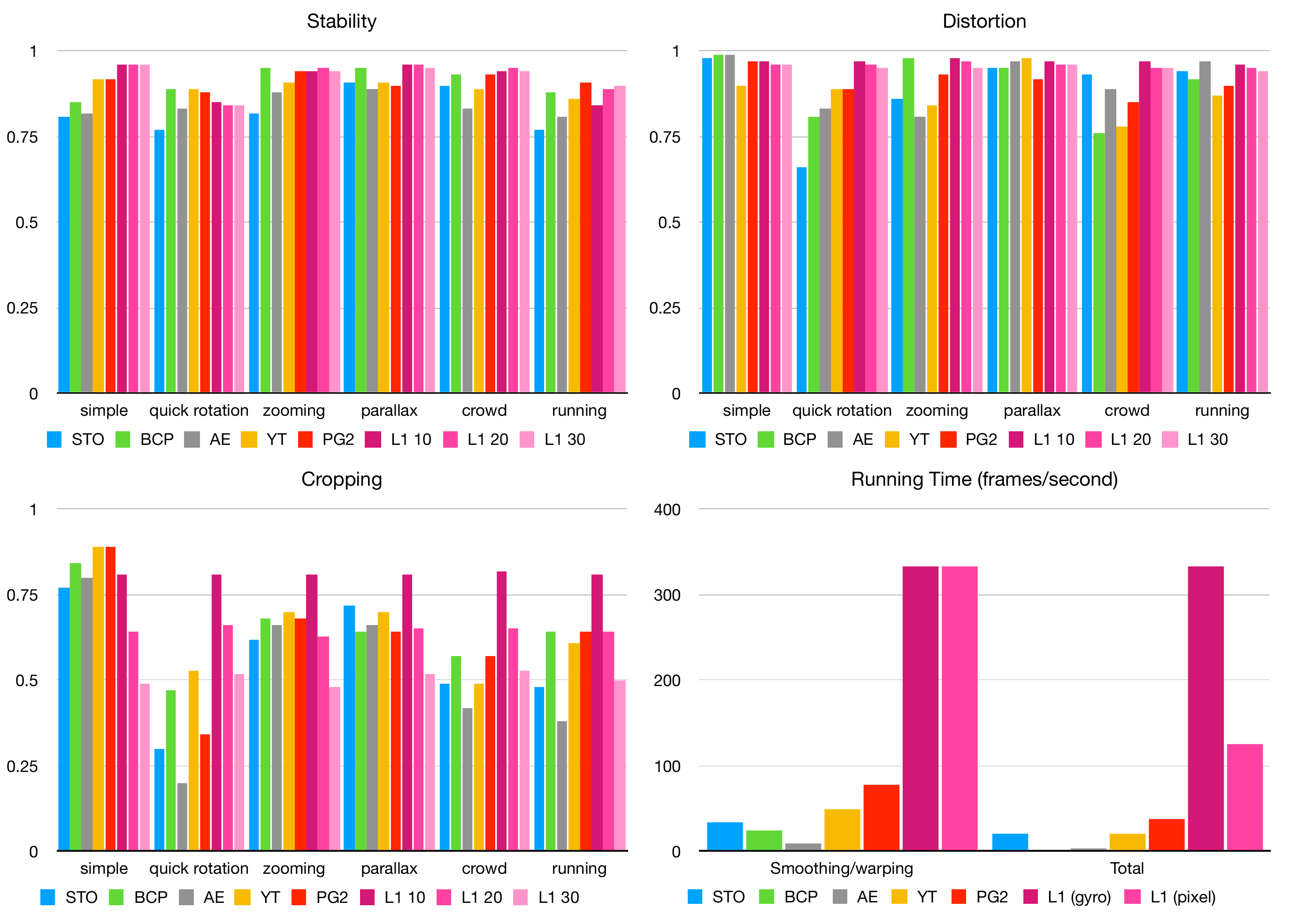}
\end{center}
\caption{\small{\emph{Stability}, \emph{distortion}, \emph{cropping}, and \emph{running time} metrics proposed in \cite{zhang2017geodesic}. Comparisons of our method (L1), with a 10\%, 20\%, or 30\% crop budget, against `spatially and temporally optimized' (STO) \cite{wang2013spatially}, `bundled camera paths' (BCP) \cite{liu2013bundled}, `geodesic video stabilization' (homography version) (PG2) \cite{zhang2017geodesic}, Adobe After Effect (AE) warp stabilizer (based on \cite{liu2011subspace}), and YouTube stabilizer (based on \cite{grundmann2011auto}). We use the dataset provided in \cite{liu2013bundled} consisting of 174 videos divided into several categories. For \emph{running time}, we consider both the total time and smoothing/warping time: L1 total time depends on the input motion analysis method: pixel-based analysis increases the total time while inertial sensor analysis (gyro) adds negligible time.}}
\label{fig:zhang_metrics}
\end{figure}

\textbf{Quantitative comparisons:} We follow the quantitative evaluation strategy proposed by \cite{zhang2017geodesic}, comparing \emph{stability}, \emph{distortion}, \emph{cropping ratio}, and \emph{running time} metrics as defined in~\cite{zhang2017geodesic} against five representative stabilization methods. Results are shown in figure \ref{fig:zhang_metrics} and details are in supplement. Our method has by far the best \emph{running time} and is competitive in the other metrics across all categories.
\\
\textbf{Qualitative comparisons:} 
We provide a collection of L1-stabilized videos in the supplementary material; details are in supplement.
One of our main contributions is solving with homographies (versus affinities as used in~\cite{grundmann2011auto}), which allow for keystone (perspective) correction. Videos described in supplement demonstrate the superiority of homographies over affinities in removing residual keystone wobble.
Next, we provide specific comparisons to ~\cite{grundmann2011auto}, detailed in supplement. Since~\cite{grundmann2011auto} refine the affine solution with homographies in a `residual motion suppression' step, they do not suffer from the severe residual keystone discussed above, but the post-processing step can consume significant additional crop on top of the fraction allowed in the optimization.
Finally, supplement discusses stabilization of very shaky videos like \emph{wakatipu-l1-affine.mp4}. The local nature of the log approximation might raise concern when aggressive stabilization is required, but since we control the correction magnitude through the fidelity objective and crop and distortion constraints, we avoid making corrections severe enough to invalidate the approximation. The supplement also includes examples showing the effects of windowing.
\\
\textbf{Ablation study:} An ablation study in supplement isolates the effects of individual objective terms and constraints. We evaluate the effect of the derivative terms, compare different crop allowances, and specifically assess key innovations relative to~\cite{grundmann2011auto}. 
\\
\textbf{Performance:} Timing data is shown in Figure \ref{fig:wind_l1_time_mem}. Our methods runs at about 300fps on an iPhone XS (assuming inertial sensor measurements for the input path). With windowing, runtime scales linearly in the number of frames. In a table in the supplement, we compare our performance with that of several alternatives (a superset of those in figure \ref{fig:zhang_metrics}); our method is orders of magnitude faster than the 2.5-3D methods, and only comparable in performance to 2D methods; of the 2D methods our goals are most similar to \cite{grundmann2011auto}'s.

\section{Conclusions, Limitations, and Future Work} 

Building on ~\cite{grundmann2011auto}, we develop a convex optimization-based video stabilization framework that can create stationary, panning, and constant-acceleration motions characteristic of professionally shot video, accurately satisfy a crop constraint, stay close to the original camera path, and handle salient objects. Our contributions include a log-homography transform enabling homography modeling without loss of convexity, a principled linearization strategy for crop and saliency constraints, quadratic objective terms to encourage fidelity to the input path and possibly center salient objects, and a windowing strategy that approximates the solution in linear time and bounded memory.
Limitations of this method include its 2D nature, which cannot handle depth/parallax but is essential for speed, and the fact that the model weights must currently be hand-tuned. In future work, we would like to study data-driven tuning of the model parameters using professionally-shot versus handheld videos as labeled data for supervised learning.

{\small
\bibliographystyle{ieee_fullname}
\bibliography{../l1_stab_bibliography}
}

\end{document}